\documentclass[11pt,a4paper]{article}
\pdfoutput=1
\usepackage{authblk}
\usepackage[nohyperref]{acl2017}
\usepackage{times}
\usepackage{latexsym}
\usepackage{url}
\usepackage{graphicx} 
\usepackage{amsmath,amssymb}
\usepackage{ifthen}
\usepackage{color}
\usepackage{pgfplots}

\usepackage{subcaption}
\usepackage[font={footnotesize}]{caption}

\aclfinalcopy 

\newcommand\refsec[1]{Section~\ref{sec:#1}}

\newcommand\reffig[1]{Figure~\ref{fig:#1}}

\ifthenelse{\isundefined{\definition}}{\newtheorem{definition}{Definition}}{}
\ifthenelse{\isundefined{\assumption}}{\newtheorem{assumption}{Assumption}}{}
\ifthenelse{\isundefined{\proposition}}{\newtheorem{proposition}{Proposition}}{}
\ifthenelse{\isundefined{\theorem}}{\newtheorem{theorem}{Theorem}}{}
\ifthenelse{\isundefined{\lemma}}{\newtheorem{lemma}{Lemma}}{}
\ifthenelse{\isundefined{\corollary}}{\newtheorem{corollary}{Corollary}}{}
\ifthenelse{\isundefined{\alg}}{\newtheorem{alg}{Algorithm}}{}
\ifthenelse{\isundefined{\example}}{\newtheorem{example}{Example}}{}

\newcommand\nl[1]{{\it``#1''}} 

\newcommand\zl[1]{\text{\footnotesize{\tt #1}}} 

\newcommand\reverse[1]{\mathbf R[#1]}

\title{Neural Semantic Parsing over Multiple Knowledge-bases}

\author[1,2]{\textbf{Jonathan Herzig}}
\author[1]{\textbf{Jonathan Berant}}
\affil[1]{Tel-Aviv University, Tel Aviv-Yafo, Israel}
\affil[2]{IBM Research, Haifa 31905, Israel}
\affil[ ]{\tt {jherzig@gmail.com, joberant@cs.tau.ac.i}}

\date{}

\begin{document}
\maketitle
\begin{abstract}
  A fundamental challenge in developing semantic parsers is the paucity of strong
  supervision in the form of language utterances annotated with logical
  form. In this paper, we propose to exploit structural regularities in language
  in
  different domains, and train semantic parsers over multiple knowledge-bases
  (KBs), while sharing information
  across datasets. We find that we can substantially improve parsing
  accuracy by training a single sequence-to-sequence model over multiple KBs,
  when providing an encoding of the domain at decoding time. Our model
  achieves state-of-the-art performance on the \textsc{Overnight} dataset
  (containing eight domains), improves performance over a single KB baseline
  from 75.6\% to 79.6\%, while obtaining a 7x reduction in the number of model
  parameters.
\end{abstract}

\section{Introduction}
\label{introduction}

Semantic parsing is concerned with translating language utterances into executable
logical forms and constitutes a key technology for developing conversational
interfaces
\cite{zelle96geoquery,zettlemoyer05ccg,kwiatkowski11lex,liang11dcs,artzi2013weakly,berant2015agenda}.

A fundamental obstacle to widespread use of semantic parsers is the high
cost of annotating logical forms in new domains. To tackle this
problem, prior work suggested strategies such as training from denotations
\cite{clarke10world,liang11dcs,artzi2013weakly}, from
paraphrases \cite{berant2014paraphrasing,wang2015overnight} and from declarative
sentences \cite{krishnamurthy2012weakly,reddy2014large}.

\begin{figure}[t]
  {\footnotesize
	\setlength{\fboxsep}{8pt}
	\fbox{
		\parbox{\columnwidth}{
		
		\textbf{Example 1}
		
    Domain: \textsc{Housing}
		
    \nl{Find a housing that is \textbf{no more than} $800$ square feet.},

    \zl{Type.HousingUnit $\sqcap$ Size.$\leq.800$}

    Domain: \textsc{Publications}
		
		\nl{Find an article with \textbf{no more than} two authors}
		
    $\zl{Type.Article} \sqcap \reverse{\lambda x.\text{count}(\zl{AuthorOf}.x$)}.$\leq.2$
  	\vspace{0.1cm}
		
		\textbf{Example 2}
		
    Domain: \textsc{Restaurants}
		
    \nl{which restaurant has \textbf{the most} ratings?}
		
    $\text{argmax}(\zl{Type.Restaurant},\reverse{\lambda
    x.\text{count}(\reverse{\zl{Rating}}.x)})$
		
    Domain: \textsc{Calendar}
		
    \nl{which meeting is attended by \textbf{the most} people?}
		
    $\text{argmax}(\zl{Type.Meeting},\reverse{\lambda
    x.\text{count}(\reverse{\zl{Attendee}}.x)})$
		}}
  }
  \caption{Examples for natural language utterances with logical forms in
    lambda-DCS \cite{liang2013lambdadcs} in
  different domains that share structural regularity (a
comparative structure in the first example and a superlative in the second). }
    
	\label{fig:shared}
\end{figure}

In this paper, we suggest an orthogonal solution:
to pool examples from multiple datasets in different domains, each corresponding to a
separate knowledge-base (KB), and train a model over all examples. 
This is motivated by an observation that while KBs differ in their entities and properties, 
the structure of language composition repeats across domains
(Figure~\ref{fig:shared}). E.g., a superlative in language will correspond to an `argmax', and
a verb followed by a noun often denotes a join operation. 
A model that shares information across domains can improve generalization
compared to a model that is trained on a single domain only. 

Recently, \newcite{jia2016recombination} and \newcite{dong2016logical} 
proposed sequence-to-sequence models for semantic parsing. Such neural models
substantially facilitate information sharing, as both
language and logical form are represented with similar abstract vector
representations in all domains. We build on their work and examine models that
share representations across domains during encoding of language and decoding of
logical form, inspired by work on domain adaptation
\cite{daume07easyadapt} and multi-task
learning
\cite{caruana97multitask,collobert11scratch,luong2016iclr_multi,firat2016multi,johnson2016google}.
We find that by providing the decoder with a representation of the domain, we can 
train a single model over multiple domains and substantially improve accuracy
compared to models trained on each domain separately.
On the \textsc{Overnight} dataset, this improves accuracy from 75.6\% to
79.6\%, setting a new state-of-the-art, while
reducing the number of parameters by a factor of 7. To our knowledge, this work
is the first to train a semantic parser over multiple KBs.

\section{Problem Setup}
\label{sec:models_basic}

We briefly review the model presented by \newcite{jia2016recombination},
which we base our model on.

Semantic parsing can be viewed as a sequence-to-sequence problem
\cite{sutskever2014sequence}, where a
sequence of input language tokens $x = x_1, \dots, x_m$ 
is mapped to a sequence of output logical tokens $y_1,\dots,y_n$
.

The \textbf{encoder} converts $x_1, \dots, x_m$ into a sequence of context
sensitive embeddings $b_1,\dots,b_m$ using a bidirectional
RNN~\cite{bahdanau2015neural}: a forward RNN generates hidden
states $h^F_1,\dots,h^F_m$ by applying the LSTM recurrence
~\cite{hochreiter1997lstm}: $h^F_i = LSTM(\phi^{(in)}(x_i), h^F_{i-1})$,
where $\phi^{(in)}$ is an embedding function mapping a word $x_i$ to a
fixed-dimensional vector. A backward RNN similarly generates hidden states
$h^B_m,\dots,h^B_1$ by processing the input sequence in reverse. Finally, for
each input position $i$, the representation $b_i$ is the concatenation $[h^F_i, h^B_i]$ .
An attention-based
\textbf{decoder}~\cite{bahdanau2015neural,luong2015translation} generates output
tokens one at a time. At each time step $j$, it generates $y_j$ based on the current hidden state $s_j$, then updates the hidden state $s_{j+1}$ based on $s_j$ and $y_j$. Formally, the decoder is defined by the following equations:
\begin{equation}
  \begin{split}
  s_1 &= \tanh(W^{(s)}[h^F_m, h^B_1 ]) , \\
    e_{ji} &= s_j^{\top}W^{(a)}b_i , \\
  \alpha_{ji} &= \frac{\exp(e_{ji})}{\sum_{i'=1}^m e_{ji'}} , \\
  c_j &= \sum_{i=1}^{m} \alpha_{ji}b_i , \\
  p(y_j &= w \mid x, y_{1:j-1}) \propto \exp(U[s_j, c_j]) , \\
  s_{j+1} &= LSTM([\phi^{(out)}(y_j),c_j],s_j) ,
  \end{split}
\label{eq:dec_lstm}
\end{equation} 
where $i \in \{1,\dots,m\}$ and $j \in \{1, \dots, n\}$. The matrices $W^{(s)}$,
$W^{(a)}$, $U$, and the embedding function $\phi^{(out)}$ are decoder parameters. We also employ attention-based
copying as described by \newcite{jia2016recombination}, but omit details for
brevity.

The entire model is trained end-to-end by maximizing $p(y \mid x) =
\prod_{j=1}^n p(y_j \mid x, y_{1:j-1})$.

\section{Models over Multiple KBs}
\label{sec:models_da}

In this paper, we focus on a setting where we have access to $K$ training sets
from different domains, and each domain corresponds to a different KB.
In all domains the input is a language utterance and the label is a
logical form (we assume annotated logical forms can be
converted to a single formal language such as lambda-DCS in~\reffig{shared}).
While the mapping from words
to KB constants is specific to each domain, we expect that the manner in which
language expresses composition of meaning to be shared across domains. We
now describe architectures that share information between the encoders and
decoders of different domains.

\newsavebox{\largestimage}
\begin{figure*}%
\centering
\savebox{\largestimage}{\includegraphics[width=\columnwidth]{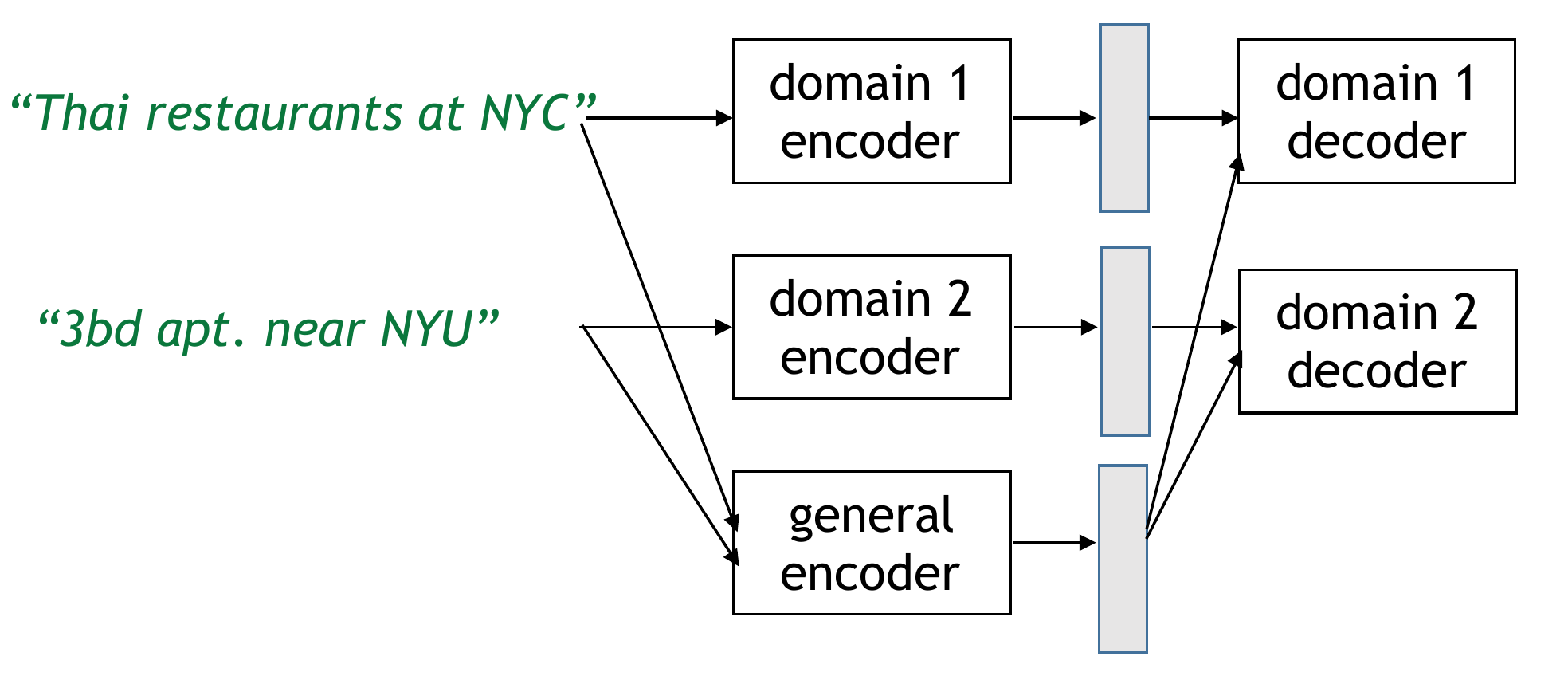}}
\begin{subfigure}[t]{1.0\columnwidth}
\raisebox{\dimexpr.5\ht\largestimage-.5\height}{
\includegraphics[width=\columnwidth]{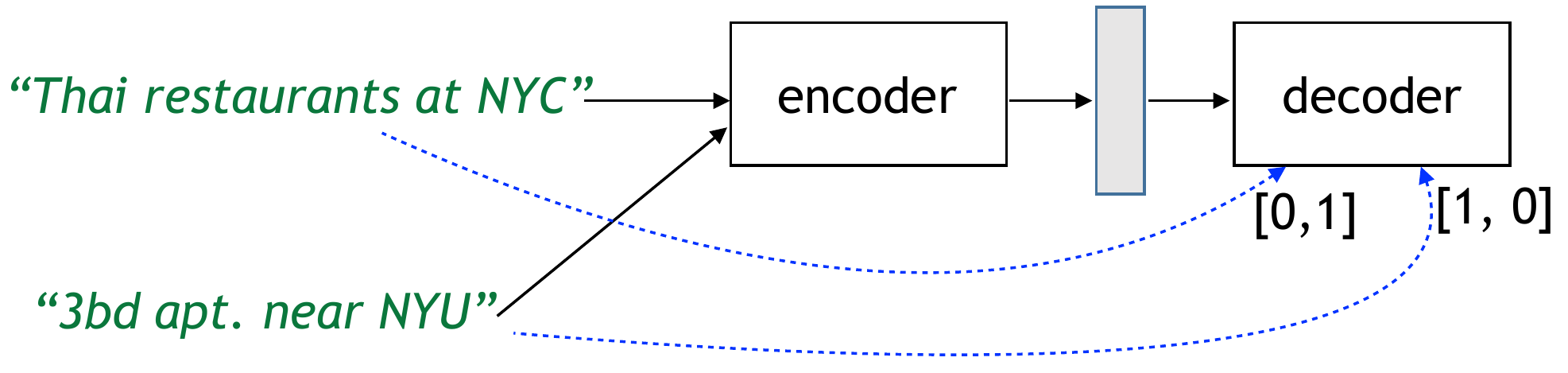}}
\end{subfigure}\hfill%
\begin{subfigure}[t]{1.0\columnwidth}
\usebox{\largestimage}
\end{subfigure}
\caption{Illustration of models with one example from the \textsc{Restaurants}
domain and another from the \textsc{Housing} domain. Left: One-to-one model with optional domain encoding. Right: many-to-many model.}
\label{fig:da_models}
\end{figure*}

\subsection{One-to-one model}
\label{sec:o2o}
This model is similar to the baseline model described in \refsec{models_basic}.
As illustrated in~\reffig{da_models}, it consists of a single encoder and a single
decoder, which are used to generate outputs for all domains. Thus, 
all model parameters are shared across domains, and the model is trained from all examples.
Note that the number of parameters does not depend on the number of domains $K$. 

Since there is no explicit representation of the domain
that is being decoded, the model must learn to identify the domain given only the input. 
To alleviate that, we encode the $k$'th domain by a one-hot vector $d_k \in
\mathbb{R}^{K}$. At each step, the decoder updates the hidden
state conditioned on the domain's one-hot vector, as well as on the previous
hidden state, the output token and the context. Formally, for domain $k$,
Equation~\ref{eq:dec_lstm} is changed:\footnote{For simplicity, we omit the
  domain index $k$ from our notation whenever it can be inferred from context.}
\begin{equation}
s_{j+1} = LSTM([\phi^{(out)}(y_j), c_j, d_k], s_j).
\end{equation}

Recently~\newcite{johnson2016google} used a similar intuition for neural
machine translation, where they added an artificial token at the beginning of each source
sentence to specify the target language. We implemented their approach and
compare to it in~\refsec{experiments}.

Since we have one decoder for multiple domains, tokens which are not in the
domain vocabulary could possibly be generated. 
We prevent that at test time by excluding out-of-domain tokens before the
softmax  ($p(y_j \mid x, y_{1:j-1})$) takes place.

\subsection{Many-to-many model}
\label{sec:m2m}
In this model, we keep a separate encoder and decoder for every domain, but augment the model with an additional encoder that
consumes examples from all domains (see~\reffig{da_models}). 
This is motivated by prior work on domain
adaptation~\cite{daume07easyadapt,blitzer2011domain},
where each example has a representation that captures domain-specific aspects of
the example and a representation that captures domain-general aspects. In our
case, this is achieved by encoding examples with a domain-specific encoder
as well as a
domain-general encoder, and passing both representations to the decoder.

Formally, we now have $K+1$ encoders and $K$ decoders, and denote by $h^{F,k}_i,
h^{B,k}_i, b^{k}_i$ the forward state, backward state and their concatenation at
position $i$ (the
domain-general encoder has index $K+1$).
The hidden state of the decoder in domain $k$ is initialized from the
domain-specific and domain-general encoder:
\begin{equation*}
  s_1 = \tanh(W^{(s)}[h^{F,k}_m, h^{B,k}_1, h^{F,K+1}_m, h^{B,K+1}_1]).
\end{equation*} 

Then, we compute unnormalized attention scores based on both encoders, and
represent the language context with both domain-general and domain-specific
representations. Equation~\ref{eq:dec_lstm} for domain $k$ is changed as follows:
\begin{align*}
  e_{ji} &= s_j^{\top}W^{(a)}[b^k_i, b^{K+1}_i], \\ 
  c_j &= \sum_{i=1}^{m} \alpha_{ji}[b^{k}_i, b^{K+1}_i]. 
\end{align*} 
In this model, the number of encoding parameters grows by a factor of
$\frac{1}{k}$, and the number of decoding parameters grows by less than a factor
of 2.

\subsection{One-to-many model}
\label{sec:o2m}
Here, a single encoder is shared, while we keep a separate
decoder for each domain. The shared encoder captures the
fact that the input in each domain is a sequence of English words. The domain-specific
decoders learn to output tokens from the right domain vocabulary.

\section{Experiments}
\label{sec:experiments}

\subsection{Data}
We evaluated our system on the \textsc{Overnight} semantic parsing dataset,
which contains $13,682$ examples of language utterances paired with logical forms across eight domains. 
\textsc{Overnight} was constructed by 
generating logical forms from a grammar and annotating them with
language through crowdsourcing.
We evaluated on the same train/test split as \newcite{wang2015overnight}, using
the same accuracy metric, that is, the proportion of questions for
which the denotations of the predicted and gold logical forms are equal.

\begin{table*}[t]
\resizebox{1.0\textwidth}{!}{
\begin{tabular}{l|cccccccc|c|c}
\hline\hline
Model    & Basketball    & Blocks        & Calendar      & Housing       & Publications  & Recipes       & Restaurants   & Social        & Avg.          & \# Model Params \\\hline
\textsc{Indep} & 85.2          & 61.2          & 77.4          & 67.7          & 74.5          & 79.2          & 79.5          & 80.2          & 75.6          & 14.1 M                                                     \\
\textsc{Many2Many}      & 83.9          & 63.2          & 79.8          & 75.1          & 75.2          & 81.5          & 79.8          & \textbf{82.4} & 77.6          & 22.8 M                                                     \\
\textsc{One2Many}      & 84.4          & 59.1          & 79.8          & 74.6          & 80.1          & 81.5          & 80.7          & 81.1          & 77.7          & 8.6 M                                                      \\
\textsc{InputToken}    & 85.9          & 63.2          & 79.2          & 77.8          & 75.8          & 80.6          & \textbf{82.5} & 81.0          & 78.2          & 2 M                                                        \\
\textsc{One2One}      & 84.9          & \textbf{63.4} & 75.6          & 76.7          & 78.9          & \textbf{83.8} & 81.3          & 81.4          & 78.3          & 2  M                                                        \\
\textsc{DomainEncoding}    & \textbf{86.2} & 62.7          & \textbf{82.1} & \textbf{78.3} & \textbf{80.7} & 82.9          & 82.2          & 81.7          & \textbf{79.6} & 2 M  
\end{tabular}}
\caption{Test accuracy for all models on all domains, along with the number of parameters for each model. 
}
\label{tab:res}
\end{table*}

\subsection{Implementation Details}
We replicate the experimental setup of~\newcite{jia2016recombination}:
We used the same hyper-parameters without tuning; we used 200 hidden
units and 100-dimensional word vectors; we initialized parameters uniformly within the interval $[-0.1, 0.1]$, and
maximized the log likelihood of the correct logical form with stochastic
gradient descent. We trained the model for 30 epochs with an initial learning
rate of 0.1, and halved the learning rate every 5 epochs, starting from epoch 15.
We replaced word vectors for words that occur only once in the training set with a universal $<$\texttt{unk}$>$ word vector.
At test time, we used beam search with beam size 5. 
We then picked the highest-scoring logical form that does not yield an executor error when its denotation is computed.
Our models were implemented in Theano~\cite{bergstra2010theano}. 

\subsection{Results}
For our main result, we trained on all eight domains all models described in Section
\ref{sec:models_da}: \textsc{One2One},
\textsc{DomainEncoding} and \textsc{InputToken} representing respectively the
basic one-to-one model, with extensions of one-hot domain encoding or an
extra input token, 
as described in Section \ref{sec:o2o}. 
\textsc{Many2Many} and \textsc{One2Many} are the models described in Sections
\ref{sec:m2m} and \ref{sec:o2m}, respectively.
\textsc{Indep} is the baseline sequence-to-sequence model described in
Section~\ref{sec:models_basic}, which trained independently on each domain.

Results show (Table~\ref{tab:res}) that training on multiple KBs improves
average accuracy over all domains for all our proposed models,
and that performance improves as more parameters are shared.
Our strongest results come when parameter sharing is maximal (i.e., single
encoder and single decoder), coupled with a one-hot domain representation at
decoding time (\textsc{DomainEncoding}). In this case accuracy improves not only
on average, but also for each domain separately. Moreover, the number of model
parameters necessary for training the model is reduced by a factor of 7.

Our baseline, \textsc{Indep}, is a reimplementation of the
\textsc{NoRecombination} model described in~\newcite{jia2016recombination},
which achieved average accuracy of $75.8\%$ (corresponds to our $75.6\%$ result).
~\newcite{jia2016recombination} also introduced a framework for generating new
training examples in a single domain through \emph{recombination}. Their model
that uses the most training data achieved state-of-the-art average accuracy of $77.5\%$
on \textsc{Overnight}. We show that by training over multiple KBs we can achieve
higher average accuracy, and our best model, \textsc{DomainEncoding}, sets a new state-of-the-art average 
accuracy of $79.6\%$.        

\begin{figure}[t]
\begin{tikzpicture}
\begin{axis}[
	width = 1.0\columnwidth,
    xlabel={Training data fraction},
    ylabel={Average accuracy},
    xmin=0, xmax=1,
    ymin=30, ymax=85,
    xtick={0, 0.1, 0.2, 0.5, 1.0},
    legend pos=south east,
    legend style={font=\scriptsize},
    ymajorgrids=true,
    grid style=dashed,
]

\addplot
	[
    color=blue!80!black,
    mark=-,
    ]
    coordinates {
    (0.1,35.50180746)(0.2,53.03196031)(0.5,70.15636033)(1,75.6)
    };
    \addlegendentry{\textsc{Indep}}
\addplot
	[
    color=orange!80!black,
    mark=o,
    ]
    coordinates {
    (0.1,33.59394876)(0.2,54.00427377)(0.5,70.32052371)(1,77.6)
    };
    \addlegendentry{\textsc{Many2Many}}
\addplot
	[
    color=teal!80!black,
    mark=triangle,
    ]
    coordinates {
    (0.1,35.97962352)(0.2,54.89484994)(0.5,70.99192567)(1,77.7)
    };
    \addlegendentry{\textsc{One2Many}}
\addplot
	[
    color=black,
    mark=diamond,
    ]
    coordinates {
    (0.1,51.96333133)(0.2,63.26885736)(0.5,74.38725608)(	1,78.2)
    };
    \addlegendentry{\textsc{InputToken}}
\addplot
	[
    color=yellow!80!black,
    mark=+,
    ]
    coordinates {
    (0.1,53.56004058)(0.2,65.63758172)(0.5,74.58613055)(1,78.3) 
    };
    \addlegendentry{\textsc{One2One}} 
\addplot
	[
    color=green!60!black,
    mark=x,
    ]
    coordinates {
    (0.1,51.54064653)(0.2,65.07475173)(0.5,75.13332641)(1,79.6)
    };
    \addlegendentry{\textsc{DomainEncoding}}

\end{axis}
\end{tikzpicture}
\caption{Learning curves for all models on the test set.}
\label{fig:learn_curve}
\end{figure}
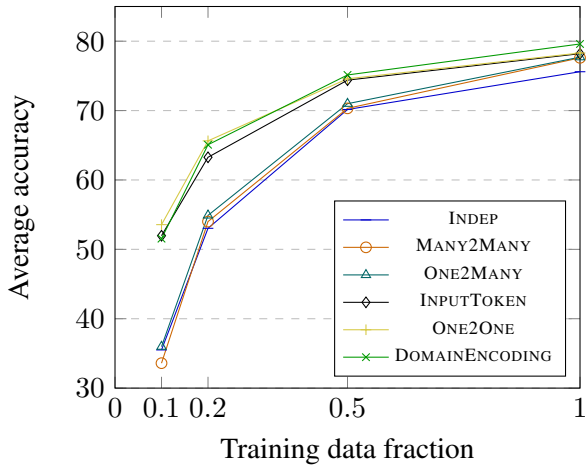

Figure~\ref{fig:learn_curve} shows a learning curve for all models on the test
set, when training on a fraction of the training data. 
We observe that the difference between models that share parameters
(\textsc{InputToken}, \textsc{One2One} and \textsc{DomainEncoding}) and models
that keep most of the parameters separate (\textsc{Indep}, \textsc{Many2Many}
and \textsc{One2Many}) is especially pronounced when the amount of data is small, reaching
a difference of more than 15 accuracy point with 10\% of the training data.
This highlights the importance of using additional data from a similar
distribution without increasing the number of parameters when there is little
data. The learning curve also suggests that the \textsc{Many2Many} model
improves considerably as the amount of data increases, and it would be
interesting to examine its performance on larger datasets.

\subsection{Analysis}
Learning a semantic parser involves mapping language phrases to KB
constants, as well as learning how language composition corresponds to logical
form composition. We hypothesized that the main benefit of training on multiple
KBs lies in learning about compositionality. To verify that, we append the
domain index to the name of every constant in every KB, and 
therefore constant names are disjoint across datasets. We train \textsc{DomainEncoding}
on this dataset and obtain an accuracy of $79.1\%$ (comparing to $79.6\%$),
which hints that most of the gain is attributed to compositionality rather than
mapping of language to KB constants.

We also inspected cases where \textsc{DomainEncoding} performed better than
\textsc{Indep}, by analyzing errors on a development set (20\% of the training
data). We found 45 cases where \textsc{Indep} makes an error (and
\textsc{DomainEncoding} does not) by predicting a wrong comparative or
superlative structure (e.g., $>$ instead of $\geq$). However, the opposite case
occurs only 29 times. This re-iterates how we learn structural linguistic
regularities when sharing parameters.  

Lastly, we observed that the domain's training set size negatively correlates
with its relative improvement in performance (\textsc{DomainEncoding} accuracy
compared to \textsc{Indep}), where Spearman's $\rho=-0.86$. This could be explained by the tendency of smaller domains to cover a smaller fraction of structural regularities in language, thus, they gain more by sharing information.

\section{Conclusion}
\label{conclusion}

In this paper we address the challenge of obtaining training data for 
semantic parsing from a new perspective. We propose that one can improve
parsing accuracy
by training models over multiple KBs and demonstrate this on the eight domains
of the \textsc{Overnight} dataset.

In future work, we would like to further reduce the burden of data gathering by
training character-level models that learn to map language phrases to KB
constants across datasets, and by pre-training language side models that 
improve the encoder from data that is independent of the KB. We also plan to
apply this method on datasets where only denotations are provided rather than
logical forms.

\section*{Reproducibility} 
All code, data, and experiments for this paper are available on the CodaLab platform
at \url{https://worksheets.codalab.org/worksheets/0xdec998f58deb4829aba80fbf49f69236/}.

\section*{Acknowledgments}
We thank Shimi Salant and the anonymous reviewers for their constructive feedback. This work was partially supported by the Israel Science Foundation, grant 942/16.

\bibliography{all}
\bibliographystyle{acl_natbib}

\end{document}